\documentclass[letterpaper, 10 pt, conference]{ieeeconf}  

\IEEEoverridecommandlockouts                              

\usepackage{graphicx}
\usepackage{hyperref}
\usepackage{tikz}
\usepackage[ruled,vlined]{algorithm2e}
\LinesNumbered
\SetArgSty{textup}
\hypersetup{
    colorlinks=true,
}
\usepackage{amsfonts}
\usepackage{amsmath}
\usepackage{cite}
\usepackage{booktabs}
\usepackage{todonotes}

\title{\LARGE \bf
TIGRIS: An Informed Sampling-based Algorithm for\\ Informative Path Planning
}

\author{ Brady Moon$^{1}$, Satrajit Chatterjee$^{1}$, and Sebastian Scherer$^{1}$
\thanks{*This work is supported by the Office of Naval Research (Grant N00014-21-1-2110). This material is based upon work supported by the National Science Foundation Graduate Research Fellowship under Grant No. DGE1745016.}
\thanks{$^{1}$Authors are with the Robotics Institute, School of Computer Science at Carnegie Mellon University, Pittsburgh, PA, USA
{\tt\small  \{bradym, satrajic, basti\}@andrew.cmu.edu}}%
}

\begin{document}

\maketitle
\thispagestyle{empty}
\pagestyle{empty}

\begin{abstract}

Informative path planning is an important and challenging problem in robotics that remains to be solved in a manner that allows for wide-spread implementation and real-world practical adoption. Among various reasons for this, one is the lack of approaches that allow for informative path planning in high-dimensional spaces and non-trivial sensor constraints. In this work we present a sampling-based approach that allows us to tackle the challenges of large and high-dimensional search spaces. This is done by performing informed sampling in the high-dimensional continuous space and incorporating potential information gain along edges in the reward estimation. This method rapidly generates a global path that maximizes information gain for the given path budget constraints. We discuss the details of our implementation for an example use case of searching for multiple objects of interest in a large search space using a fixed-wing UAV with a forward-facing camera. We compare our approach to a sampling-based planner baseline and demonstrate how our contributions allow our approach to consistently out-perform the baseline by 18.0\%. With this we thus present a practical and generalizable informative path planning framework that can be used for very large environments, limited budgets, and high dimensional search spaces, such as robots with motion constraints or high-dimensional configuration spaces. \\
\href{https://github.com/castacks/tigris}{[Code]}\footnote{ \href{https://github.com/castacks/tigris}{Codebase: https://github.com/castacks/tigris}}  
\href{https://youtu.be/bMw5nUGL5GQ}{[Video]}\footnote{ \href{https://youtu.be/bMw5nUGL5GQ}{Video: https://youtu.be/bMw5nUGL5GQ}}\\

\end{abstract}

\section{Introduction}



Real-world data gathering is an important problem being solved through autonomous robotic platforms. Some examples include spatiotemporal monitoring \cite{binney_optimizing_2013}, aerial surveillance \cite{cabreira_survey_2019,Nguyen2015LongtermIP,moon2018}, estimating urban wind fields \cite{patrikar2020}, and wildfire monitoring \cite{binneyFire}. The traditional methods when using humans to accomplish these tasks may be inefficient, infeasible, tedious, or too risky to carry out. Using autonomous robots to accomplish these tasks allows for a more efficient and safe alternative, as well as opening doors to new applications that were not formerly possible. 

\begin{figure}[th]
    \centering
    \includegraphics[trim={0cm 0cm 0cm 0cm},clip,width=.48\textwidth]{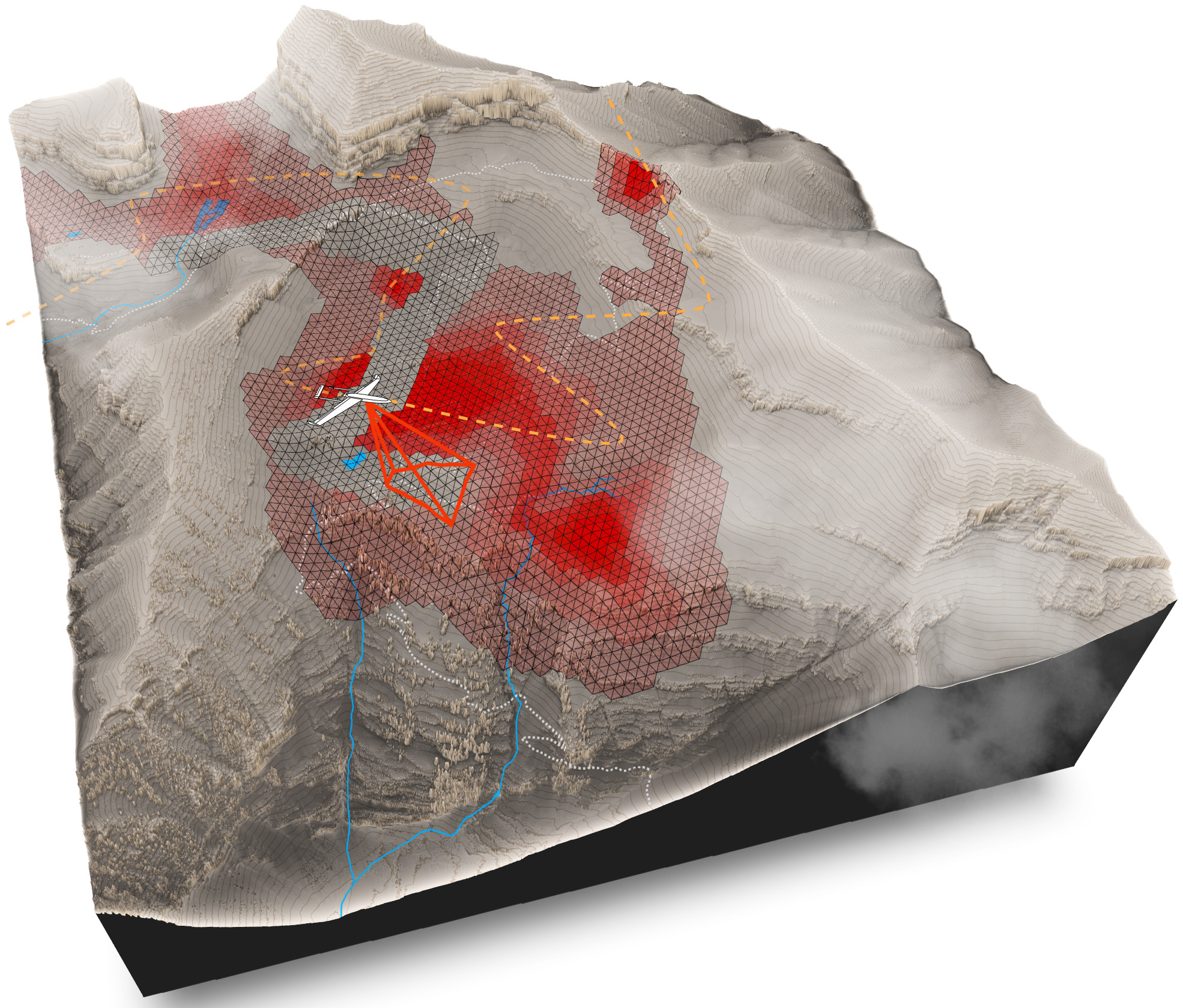} 
    \caption{An example scenario for an informative path planning problem of searching for a missing hiker. The UAV plans a path of maximum information gain over a prior belief distribution of the hiker's location.}
    \label{fig:full_overview}
\end{figure}

Each data gathering problem can present unique challenges depending on the type of data, method of collection, and sampling environment the robot must traverse, but the overall framework for solving the problem can be formulated in a unifying way. 
A typical way to formulate a data gathering problem is as an informative path planning (IPP) problem which involves finding paths that maximize information gathered while respecting budget constraints. These problems can be challenging to solve as they have been shown to be NP-hard~\cite{singh2009} and even PSPACE-hard~\cite{reif1979}.


Many approaches \cite{stanford, arora_randomized_2017, persistent_monitoring} attempt to find a global solution by discretizing the space and reducing the number of states to be as small as possible, such as planning on a grid or nodes on a graph. These solutions can be resolution complete, optimal for the given problem conditioning, but are challenging to apply to real-world scenarios and often give solutions that are clearly suboptimal for the situation. This can be due to having to reduce the size of the search space in order to get a solution quickly enough or the oversimplification of the space which results in having to ignore important factors such as data gathered between the nodes on the graph. 


The work by \cite{hollinger_sampling-based_2014} adapted common sampling-based motion planning algorithms to IPP with motion constraints and asymptotic optimality analysis. Their work quickly explores the space of possible paths to find a feasible and optimal solution and was shown to find optimal solutions faster than existing solvers. 

In this paper we present TIGRIS (Tree-based Information Gathering using Informed Sampling) where we build on the approach of \cite{hollinger_sampling-based_2014} to create a sampling-based information gathering algorithm for IPP with two key improvements:

\begin{enumerate}
    \item Perform informed sampling within a subset of the configuration space that influences the estimated information gain, allowing for planning in larger and higher dimensional search spaces by prioritizing the exploration of paths with high reward. 
    \item Include edge rewards using a novel edge cost formulation and demonstrate its advantage in spite of the extra computation. 
\end{enumerate}




We demonstrate that our algorithm works well at rapidly planning high information gain paths in high-dimensional and large spaces while being constrained to a relatively small budget relative to the size of the space. An example use-case of this algorithm is shown in Fig. \ref{fig:full_overview} where a fixed-wing UAV with a forward-facing image sensor is searching a region for a missing hiker. In such a situation, it is necessary for a path planner to be able to plan paths through areas of high potential reward in order to maximize the likelihood of collecting the most amount of relevant information while subject to a relatively small budget. 

Additionally, in this paper we also detail how we implement TIGRIS for our specific use case---searching for objects of interest using a fixed-wing UAV in an environment that is much larger than the available budget. We step through all the considerations that had to be made to successfully implement TIGRIS for this challenging problem including how to estimate edge rewards and the potential gain in information of a candidate path in the tree. We further detail our environment setup, belief representation, and how we implement our approach in a practical and feasible way. Finally, we release TIGRIS as an open source package with this paper.





The paper is organized as follows: Section \ref{sec:related} reviews the related works in this field and the contributions of our work. Section \ref{sec:problem} lays out the problem definition. Section \ref{sec:method} explains our proposed algorithm. Section \ref{sec:implementation} details our approach to implementing TIGRIS for our particular use case and all the considerations that were made in our planning environment. Section \ref{sec:results} lays out our testing framework and results compared to the baseline method. Section \ref{sec:conclusion} gives the conclusion and future work.


\section{Related Work}\label{sec:related}




Many approaches have been used for robotic data gathering in physical spaces. The IPP problem can also be formulated as an adaptive sampling problem or an orienteering problem with each having their own subtle variations in the problem formulation.

Some of the previous works have tried to solve the IPP problem using a receding horizon approach. The work by \cite{wettergreen_experimental_2016} proposed an adaptive receding horizon controller for marine monitoring that modifies the lookahead step size to not get stuck in a local optima. The work by \cite{Nguyen2015LongtermIP} compares finite horizon tree search, locally optimal mypopic search, Monte Carlo tree search, and their own versions of cluster tree search when searching for a target while gliding between thermals. They explain the strengths and weaknesses of each method and show their versions of cluster tree search perform the best in their mission framework. These receding horizon methods reactively plan locally optimal paths and do not fully leverage the known prior environment information. 

In regard to planning globally optimal paths, many approaches discretize the space and turn the problem into a graph search problem such as in \cite{singh2009, binney2010, hai2021}. However, the work by \cite{frolov2014} found that many of the graph based approaches can perform closely to that of a "lawnmower" search due to their inability to revisit areas more than once. Allowing for revisits can be very important for finding the optimal path when the information gain function is submodular or when multiple revisits are necessary for a given task. The work in \cite{arora_randomized_2017} formulates the IPP as an orienteering problem and presents an anytime solution to informative path planning in a discrete space. The authors do note that their approach only offer low run-times when the solution space has multiple paths that are near-optimal, and their approach also does not allow for revisits. The run-times become quickly intractable as the probability of near-optimal paths existing in the solution space becomes small. Discretized approaches can also start to become computationally intractable as the graph size increases \cite{binney_optimizing_2013, arora_randomized_2017} which makes them infeasible in large search spaces. 

As an alternative to this approach, information gathering problems can also be formulated in continuous spaces. Examples of this approach can be seen in the works by \cite{schmid_efficient_2020, hollinger_sampling-based_2014, hollinger_long-horizon_2015}. In all these approaches the authors adopt a sampling-based approach where points in continuous space are sampled and added to a tree. An anytime best path is chosen in the tree that maximizes information gain while being subject to a budget constraint. A drawback of these approaches is that the problem formulations can still suffer in large spaces. As the dimensionality increases, the space needed to search can also explode relative to the size of the space with high information reward. 

This raises the question: how do we perform anytime informative path planning in large high-dimensional spaces efficiently? In this work we explore this question and present a practical solution to this challenging problem. Our method uses a sampling-based approach but leverages the prior environment information to perform informed sampling within the space. This prioritizes building the tree in areas of high reward even when the dimensionality of the space is large. Our inclusion of edge rewards also helps in more accurately estimating path rewards thereby quickly expanding the tree in large spaces without sacrificing accurate path reward estimation. 






\section{Problem Definition}
\label{sec:problem}

Let $\mathcal{T}$ represent a sensor trajectory in $\mathbb{T}$ and $C(\mathcal{T})$ be the cost function for a given trajectory. Let $I(\mathcal{T})$ be the expected reward or information gain from a trajectory and $B$ be the budget constraint. 

We define the informative path planning problem as maximizing information gain subject to a budget constraint as in the following:

\begin{equation*}
    \mathcal{T}^*=\underset{\mathcal{T}\in \mathbb{T}}{\arg\max}~I(\mathcal{T}) \text{ s.t.\ $C(\mathcal{T}) \leq B$}
    \label{equ:1}
\end{equation*}

where an optimal trajectory $\mathcal{T}^*$ is found that does not exceed the budget $B$.

In this formulation, the information reward function $I(\cdot)$ is not constrained to being modular and could also be time-varying modular or even submodular \cite{krause}.



\section{Methodology}\label{sec:method}




On a high level, TIGRIS is similar to other sampling-based planners in that points are sampled within the search space and used to build out a tree. For the case of IPP, each node in the tree contains the state, path cost, and path information gain. At the end of the planning time, the node with the highest value of path information is returned. When searching over a large space while subject to a limited budget, our planner seeks to build high reward paths quickly through prioritizing the sampling of regions with high reward. This allows us to generate plans that outperform random sampling approaches throughout the planning time. The code for TIGRIS is described in Alg. \ref{alg:ours}.

Let $\mathcal{X}$ represent the state space and $\mathbf{x}_{start}$ being the starting configuration within $X_{free}$. $T$ represents the time allowed for planning and $\Delta$ is the extend distance when building the tree. For brevity, we will use the notation $X \xleftarrow{+} \{\mathbf{x}\}$ and $X \xleftarrow{-} \{\mathbf{x}\}$ to represent the operations $X \leftarrow X \cup \{\mathbf{x}\}$ and $X \leftarrow X \setminus \{\mathbf{x}\}$ respectively.

The algorithm starts by calculating the information reward for the starting configuration $\mathbf{x}_{start} \in \mathcal{X}$ with $I(\cdot)$ and sets the cost to zero (Alg. \ref{alg:ours} Line 1). It then initializes the starting node $n_{start}$, adds it to the vertices set $V$, and initializes the edges set $E$ and graph $G$ (Alg. \ref{alg:ours} Line 2-3).

While the computation time has not exceeded the allowed planning time $T$, the tree will continue to be expanded. This begins with a new configuration $\mathbf{x}_{sample}$ being sampled using $InformedSample(\cdot)$ (Alg. \ref{alg:ours} Line 5), which in our case we use weighted sampling where the probability is directly proportional to the information reward. Details on how we implemented this sampling efficiently is found in Section \ref{vehandsens}.

The nearest node $n_{nearest} \in V \setminus V_{closed}$ to $\mathbf{x}_{sample}$ is found using the function $Nearest(\cdot)$ (Alg. \ref{alg:ours} Line 6). Efficient nearest neighbors search could be implemented using structures such as a k-d tree or R-tree.

A feasible point $\mathbf{x}_{feasible}$ is found by extending from the nearest node by a maximum cost $\Delta$ using the function $Steer(\cdot)$ (Alg. \ref{alg:ours} Line 7). This feasible point may have an edge cost less than $\Delta$ because the $Steer(\cdot)$ function will extend until a collision or the budget is reached. This is a subtle difference from \cite{hollinger_sampling-based_2014} in that ours will not allow points to extend beyond the budget but will rather reduce the extended distance to create a path that uses the entire budget.

If the feasible point is not the same as the sample point, then near points $n_{near} \in V \setminus V_{closed}$ within a radius from $\mathbf{x}_{feasible}$ are found using $Near(\cdot)$ and set as $N_{near}$ (Alg. \ref{alg:ours} Line 8-9). This radius could be predefined or set using a function like in \cite{karaman2010} which defines a closed hypersphere of radius $r_n$ where $n$ is the number of nodes.

\begin{algorithm}[t]
\SetInd{0.4em}{0.8em}

$I_s \leftarrow I(\mathbf{x}_{start})$; $C \leftarrow 0$\;
$n_{start} \leftarrow \{\mathbf{x}_{start}, I_s, C\}$\;
$V \leftarrow {n_{start}}$; $E \leftarrow \emptyset$; $V_{closed} \leftarrow \emptyset$; $G \leftarrow (V,E)$\;
\While{computation time $< T$}{
$\mathbf{x}_{sample} \leftarrow InformedSample(\mathcal{X})$\;
$n_{nearest} \leftarrow Nearest(\mathbf{x}_{sample}, V \setminus V_{closed})$\;
$\mathbf{x}_{feasible} \leftarrow Steer(\mathbf{x}_{sample}, n_{nearest}, \Delta, X_{free})$\;
\If{$\mathbf{x}_{feasible} \ne n_{nearest}$}{
$N_{near} \leftarrow Near(\mathbf{x}_{feasible}, V \setminus V_{closed})$\;
\For{$n_{near} \in N_{near}$}{
$\mathbf{x}_{new},e_{new}\!\leftarrow\!Steer(\mathbf{x}_{feasible}, n_{near},\!\Delta, X_{free})$\;
\If{$\mathbf{x}_{new} \neq n_{near}$}{
$I_{new} \leftarrow I(\mathbf{x}_{new}, e_{new}, n_{near})$\;
$C_{new} \leftarrow C_{n_{near}} + Cost(e_{new})$\;
$n_{new} \leftarrow \{\mathbf{x}_{new}, I_{new}, C_{new}\}$\;
\If{not $Prune(n_{new})$}{
$E \xleftarrow{+} \{ (n_{near}, n_{new})\}$\;
$V \xleftarrow{+} \{ n_{new}\}$\;
$G \leftarrow (V,E)$\;
\If{$C_{new} = B$}{
$V_{closed} \xleftarrow{+} \{ n_{new}\}$\;
}
}
}
}
}
}
\Return $\mathcal{T} \leftarrow BestPath(G)$ 

\caption{TIGRIS($\mathbf{x}_{start} \in X_{free}$, $B$, $T$, $\Delta$, $\mathcal{X}$)\label{alg:ours}}
\end{algorithm}

For each node $n_{near}$ in the set $N_{near}$, and new state $\mathbf{x}_{new}$ and edge $e_{new}$ is extended from $n_{near}$ to $\mathbf{x}_{feasible}$ using the $Steer(\cdot)$ function of cost $\Delta$ (Alg. \ref{alg:ours} Line 11). If the new state $\mathbf{x}_{new}$ is not the same as $\mathbf{x}_{feasible}$ (meaning it is not trapped or in the closed set), then a new node $n_{new}$ will be created. The information reward for the node is estimated using $I(\cdot)$ which calculates the reward for the trajectory up to the new state $\mathbf{x}_{new}$, including the new edge $e_{new}$ and entire trajectory of edges and states preceding it. Depending on the modularity and temporal dependence of your information function, calculating the reward for a new node might be as simple as appending the additional information gain from the new state and edge or, alternatively, the entire trajectory up to that point may need to be calculated. 

After calculating the estimated information reward, the cost of the new edge is found with $Cost(\cdot)$ and appended to the cost of the near node $C_{n_{near}}$. With the new state, information, and cost, a new node is now created (Alg. \ref{alg:ours} Line 13-15). However, before adding this new node to the tree, it is first checked for if it should rather be pruned. The $Prune(\cdot)$ function examines nearby nodes to see if it is worth adding the new node. This could be a simple heuristic of seeing if a nearby node has a lower cost and higher reward, but this heuristic can have implications on finding the optimal solution. A variety of other heuristics can be used to reduce computational complexity, but, to ensure optimality, the upper bound on reward for the nodes would need to be used within the $Prune(\cdot)$ function. 

If the node isn't pruned, the edge is added to the set $E$, the node is added to the set $V$, and the graph is updated (Alg. \ref{alg:ours} Line 17-19). If the cost of the new node is equal to the budget, then the node is also add to the closed set $V_{closed}$. Lastly, once the while loop is terminated when the max planning time is exceeded, the path with the highest information reward is returned.

\section{Implementation}\label{sec:implementation}
 
We implemented TIGRIS with the objective of reducing entropy with a bias toward increasing belief probabilities. Our data gathering is performed using a fixed wing UAV and a static forward facing camera. We discretize our space into grid cells with index $i$, and let $X_i$ be the event that an object of interest lies within the cell at index $i$. Therefore $P(X_i)$ is the probability of the event, such as a lost hiker being in the cell. In the formulation of entropy reduction, the value within a cell could easily be any other metric that we want to reduce uncertainty for such as a model for radioactive strength in a space or a temperature field. We assume a prior belief over the map that we then leverage to plan a path that reduces the overall entropy. 

Let $\mathbf{x} \in \mathbb{R}^3 \times SO(2)$ represent the state of a fixed wing UAV in inertial space. We assume constant curvature and connect states using Dubins paths. The sensor is statically mounted facing forward on the UAV with a downward pitch angle of $\theta$ and the heading matching the vehicle of $\psi$ as seen in Fig. \ref{fig:sensor_config}. 

For the $Prune(\cdot)$ function, we chose an aggressive heuristic of not adding a new node if a nearby node has a lower cost and higher reward. For our nearest neighbor search we used the Open Motion Planning Library \cite{sucan2012the-open-motion-planning-library}. 
Our function for $InformedSample(\cdot)$ is seen in Alg. \ref{alg:sample}. We implemented weighted sampling with the probability of selection being proportional to information reward, however calculating the information reward for every state would be computationally infeasible. To handle this, we compute the information reward for viewing each cell at an optimal range and then use that value for a weighted sample over cells (Alg. \ref{alg:sample} Line 1). We then sample uniformly within the subset of the space that is independent of the sampled cell, $z$ and $\psi$ (Alg. \ref{alg:sample} Line 2-3). The full sampled state can then be found using $v_{opt}$, which is a fraction reflecting the desired vertical placement of the target between the bottom of the frame and the optical axis in the image frame (Alg. \ref{alg:sample} Line 4-7). 

 
 \begin{algorithm}[]
$x,y \leftarrow WeightedSampler()$\;
$z \leftarrow SampleZ()$\;
$\psi \leftarrow Sample\psi()$\;
$\Delta_{xy} \leftarrow z\tan(\theta - v_{opt}\frac{\Theta_v}{2})$\;
$x \leftarrow x - \Delta_{xy}\cos(\psi)$\;
$y \leftarrow y -  \Delta_{xy}\sin(\psi)$\;

\Return $x,y,z,\psi$ 

\caption{InformedSample()\label{alg:sample}}
\end{algorithm}

\subsection{Sensor Model}
A sensor model is necessary for belief map updates given sensor measurements. The function approximating the sensor performance could be derived through testing a given perception system and approximated with a high order polynomial or even be a lookup based on the testing data. However the method, it is important to ensure the model maps closely and accurately to the actual system, otherwise the search algorithm could lead to erroneous behaviors. There are many factors that could be included as inputs to a sensor model include target range, sensor velocity, sensor acceleration, shutter speed, sensing angle, altitude, target material type, weather, fog, lighting conditions, and target behavior.  

\begin{figure}[b]
    \centering
    \includegraphics[trim={0cm 0cm 0cm 0cm},clip,width=.49\textwidth]{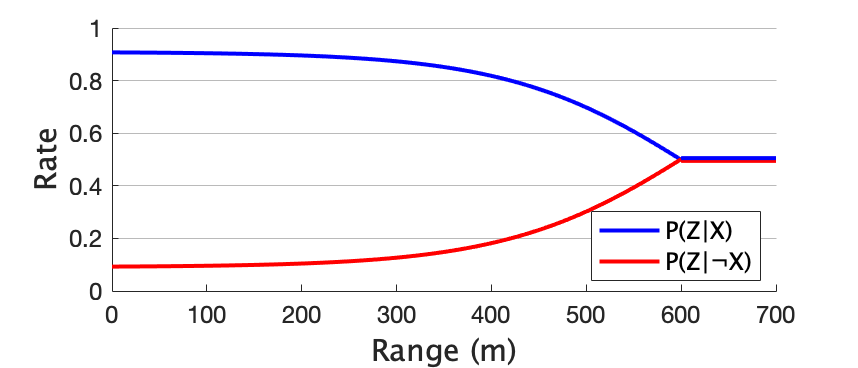} 
    \caption{An example sensor model of (\ref{equ:tpr}) with blue being the true positive rate and red being the false positive rate.}
    \label{fig:tpr}
\end{figure}

In our implementation, we are modeling the performance of a perception system that uses an electro-optical sensor. Let the variable $Z$ represent the event of a positive measurement. We model both the true positive rate $P(Z|X)$ and false positive rate $P(Z|\neg X)$ using (\ref{equ:tpr}), with the false positive rate just being the complement of the equation in our case, and $r$ being the range of the object from the sensor and $a$, $b$, and $c$ being parameters to fit the model to the performance of a perception pipeline. $\beta$ is where the sensors performance drops off to where measurements do not change the belief space. Fig.~\ref{fig:tpr} shows the plots of (\ref{equ:tpr}) and its complement.

\begin{equation}
    f(r) = \begin{cases} \frac{1}{a + e^{b(r-c)}} & r \leq \beta \\ 0.5 & r > \beta \end{cases}
    \label{equ:tpr}
\end{equation}


\subsection{Information Reward Function}

We base our information reward function on the reduction in entropy \cite{liam2014} from a new measurement. The full equation for Shannon entropy in our case is 

\begin{equation*}
    H(X) = -P(X)\log P(X)-P(\neg X) \log P(\neg X)
\end{equation*}

Using Bayes' theorem to find $P(X|Z)$, the entropy reduction due to a positive measurement would be 

\begin{equation*}
    \Delta H(X|Z) = H(X)-H(X|Z)
\end{equation*}

In order to prioritize reducing entropy in areas that increase our confidence in objects locations, we also add weights to the entropy reduction. $R_p$ is the weight for increasing probabilities and $R_n$ is for decreasing probabilities.

Ideally we would compute the expectation of the entropy reduction over the entire path, but computing this for all combinations of outcomes for each path segment would be computationally expensive. We follow the same approach as \cite{hollinger_long-horizon_2015} of using the optimistic approximation of the expected reward to compute the upper bound of the reward. If $P(X) \geq 0.5$ then we assume a positive measurement $Z$ and the converse if less than $0.5$. The resulting reward function is the following:

\begin{equation}
    R(X) = \begin{cases} R_p\Delta H(X|Z) & P(X) \geq 0.5 \\ R_n\Delta H(X|\neg Z) & P(X) < 0.5 \end{cases}
    \label{equ:reward}
\end{equation}

\subsection{Estimating Edge Rewards}\label{vehandsens}

When calculating the information reward over a path, the sensor footprint can be projected onto the surface plane to find all cells within the footprint and the reward for each cell is calculated using (\ref{equ:reward}) and summed. This step becomes more complicated when including edges in the reward calculation. Discretizing the edge into nodes would have imperfect overlaps due to a non-square sensor footprint, thus not including some of the grid cells in calculating the edge reward. To mitigate this issue, large overlaps of the sensor footprints would be needed but would thus lead to a large increase in the computation time for grid cell belief updates. Rather than choosing between the tradeoffs, we compute an approximation of the edge reward by finding all cells within the sliding sensor footprint along the edge and find the minimum distance the UAV will be to each of those cells. This ensures all cells within the edge are accounted for and each belief value is updated using a measurement when the sensor is closest to that cell.

\begin{figure}[th]
    \centering
    \includegraphics[trim={0cm 0cm 0cm 0cm},clip,width=.49\textwidth]{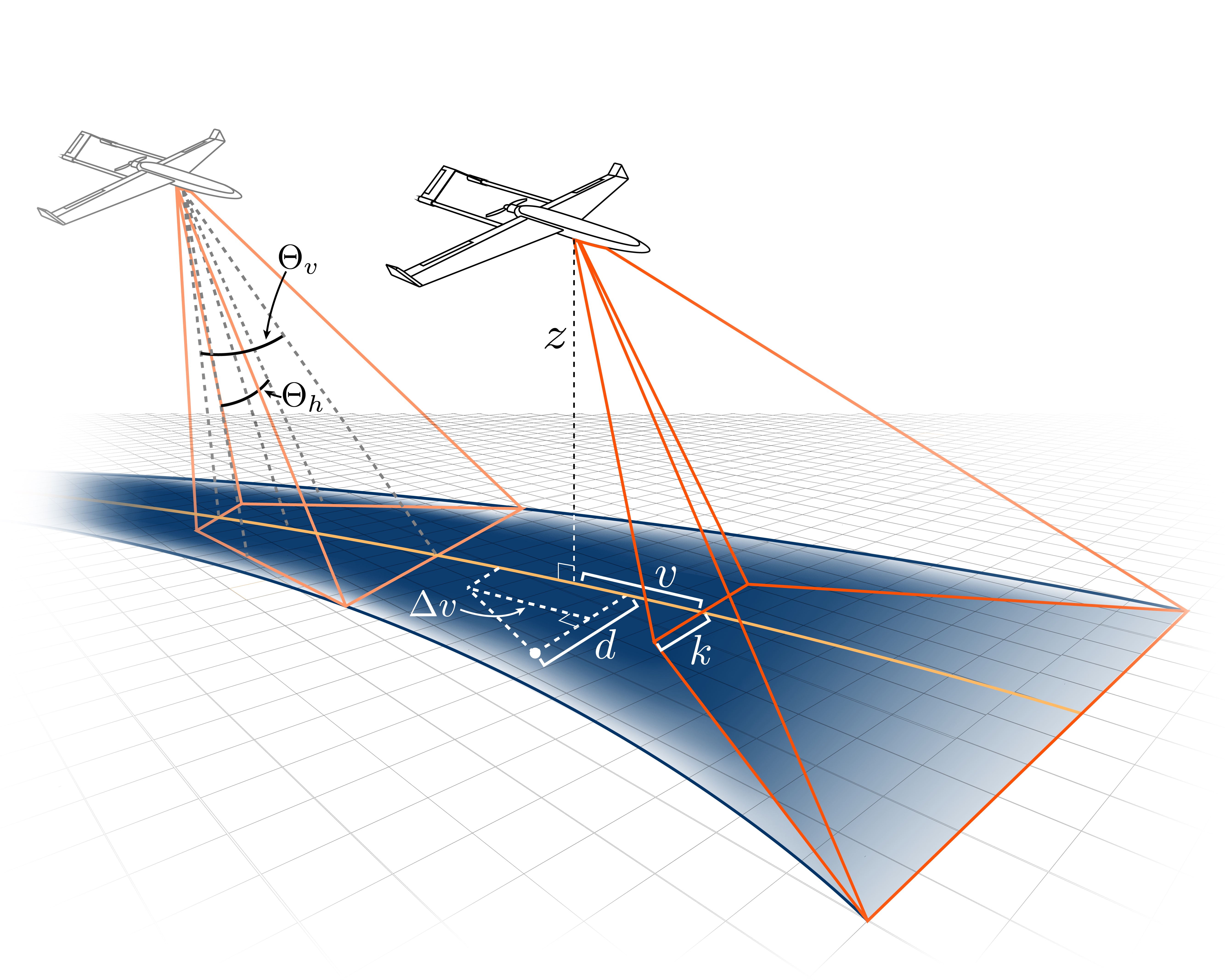} 
    \caption{A visualization of the edge reward approximation and variable definitions given a fixed forward-facing camera.
    }
    \label{fig:sensor_config}
\end{figure}

The minimum range $r_{min}$ of a point to our sensor is first found by calculating the perpendicular distance $d$ of the point to the edge line. Let $z$ be the altitude of the camera sensor above the ground, $k$ be half of the length of the bottom edge of the projected camera footprint on the ground plane, $\Theta_v=VFOV$, and $\Theta_h=HFOV$. The distance $L$ from the edge defines the distance a point will be to the sensor when it transitions from being on the bottom of the frame to the left or right edge of the image frame and is calculated using (\ref{equ:L}).

\begin{equation}
    L = \begin{cases} k & \theta > \frac{\Theta_v}{2} \\ 
                      z\tan(\frac{\Theta_h}{2}) & \theta \leq \frac{\Theta_v}{2} \end{cases}
\label{equ:L}
\end{equation}

We then calculate $r_{min}$ using (\ref{equ:edge_update}) where  $v = z\tan(\theta-\frac{\Theta_v}{2})$ and $\Delta v = \frac{d-L}{\tan(\theta)}$. The resulting $r_{min}$ is used to for each cell belief update and reward calculation of cells that are being passed over. For cells at the end of edges or at nodes, euclidean distance is used for the range. A visual representation of these calculations is seen in Fig. \ref{fig:sensor_config}.

\begin{equation}
    r_{min} = \begin{cases} \begin{aligned} \left\|\begin{bmatrix}
                                d& v&z
                                    \end{bmatrix} 
                                    \right\|_2 & ~~~\theta > \frac{\Theta_v}{2} \land d < L \\ 
                            \left\|\begin{bmatrix}d& 0& z\end{bmatrix}\right\|_2 & ~~~ \theta \leq \frac{\Theta_v}{2} \land d < L \\
                            \left\|\begin{bmatrix}d& v+\Delta v& z\end{bmatrix}\right\|_2 & ~~~ \theta > \frac{\Theta_v}{2} \land d \geq L \\
                            \left\|\begin{bmatrix}d& \Delta v & z\end{bmatrix}\right\|_2 & ~~~ \theta \leq \frac{\Theta_v}{2} \land d \geq L 
                \end{aligned}\end{cases}
\label{equ:edge_update}
\end{equation}


\section{Testing Results}\label{sec:results}

We compared our TIGRIS algorithm against the RIG-tree algorithm outlined in \cite{hollinger_sampling-based_2014} which was shown to find optimal solutions in similar testing environments. We used the same parameters for each algorithm and implemented both in C++. We also used the same information reward function when evaluating final trajectories for each algorithm. The simulation environment was a 5000x5000 meter area with grid cells of 50x50. The UAV had a budget $B=6000$ and the camera had a pitch of $\theta=65^{\circ}$. 

The simulations were run on a 3.4 GHz Intel i7 processor with 16 GB of RAM. 12000 Monte Carlo simulations were carried out with 1-12 randomly placed target belief centroids. For each centroid, a random prior mean belief value and standard deviation were selected for a Gaussian kernel. 

\begin{figure}
    \centering
    \includegraphics[trim={2.1cm 0cm 0 -.5cm 0cm},clip,width=.49\textwidth]{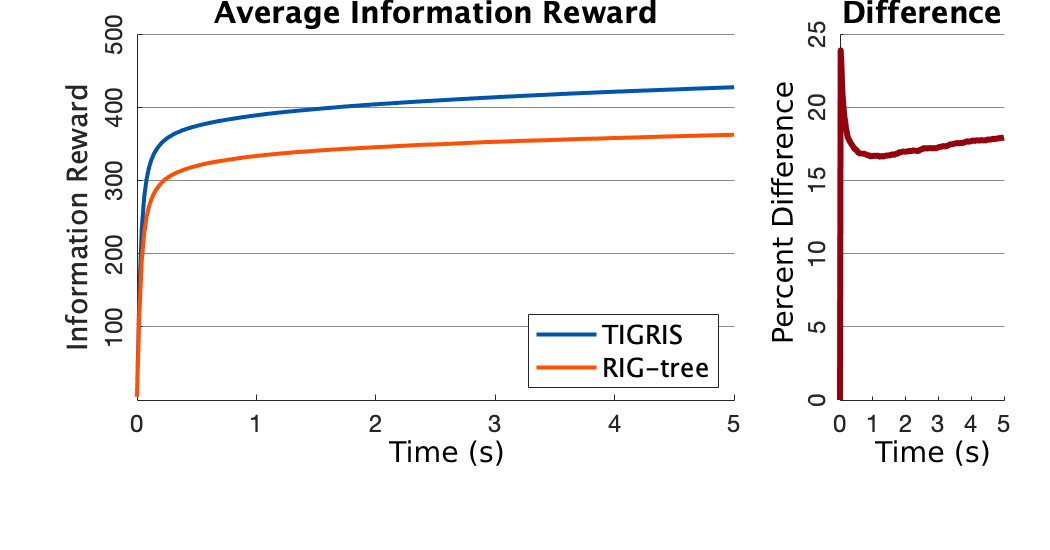}
    \caption{The results of 12000 Monte Carlo simulations. The blue line shows the average path information reward for the TIGRIS planner while the orange is for the TIG-tree. The graph on the right shows the difference in the two means.}
    \label{fig:full_test}
\end{figure}

The results for the Monte Carlo simulations are seen in Fig. \ref{fig:full_test}. The plot shows the information reward of the best path averaged over the 12000 runs. For both methods a high quality path is quickly found and both seem to be approaching an asymptotic value. It is also clear to see the improvement in TIGRIS over the entire planning time. The mean and standard deviation of information reward for TIGRIS and RIG-tree at 5 seconds is $427.53\pm126.82$ and $362.38\pm136.26$ respectively, which is an $18.0\%$ improvement. This give a P value of well less than $0.0001$ for the single tail difference between the two means. It is interesting to note the right side of Fig. \ref{fig:full_test} which shows that the percent difference in the mean rewards continues to increase at a near linear rate after the initial jump.    


Table \ref{tab:comparison} groups the results based on the number of target belief centroids that were in the testing environment. The data shows that as the density of the target belief centroids decreases the performance difference between the algorithms increases. Informed sampling becomes more of an advantage as it becomes harder for random sampling to find high-reward paths. This shows the strength of our method in sparse environments. Even at the lowest difference between the algorithms, the P-values of the difference between the mean information rewards were all far below 0.0001.


\begin{table}[b]
    \centering
    \caption{Information reward (M$\pm SD$) with decreasing sparsity in the number of target belief centroids (n) in the environment.}
    \begin{tabular}{lcccc}
    \toprule
         \textbf{Algorithm} & \textbf{n = 1-3} & \textbf{n = 4-6} & \textbf{n = 7-9} & \textbf{n = 10-12} \\
    \midrule
    TIGRIS &  \textbf{314$\pm$132} & \textbf{423$\pm$99} &  \textbf{469$\pm$86}   &  \textbf{502$\pm$78} \\
    RIG-tree \cite{hollinger_sampling-based_2014} &  235$\pm$121 & 351$\pm$108 & 411$\pm$98  &   452$\pm$91 \\
    \midrule
    Difference &  79 & 71 &  58   &  51 \\
    Percent Diff &  33.6\% & 20.5\% &  14.1\%   &  11.1\% \\

    \bottomrule
    \end{tabular}
    \label{tab:comparison}
\end{table}
\addtolength{\textheight}{-8cm}

\begin{figure}[th]
    \centering
    \includegraphics[width=.49\textwidth,trim={0cm 0cm 0cm 0cm},clip]{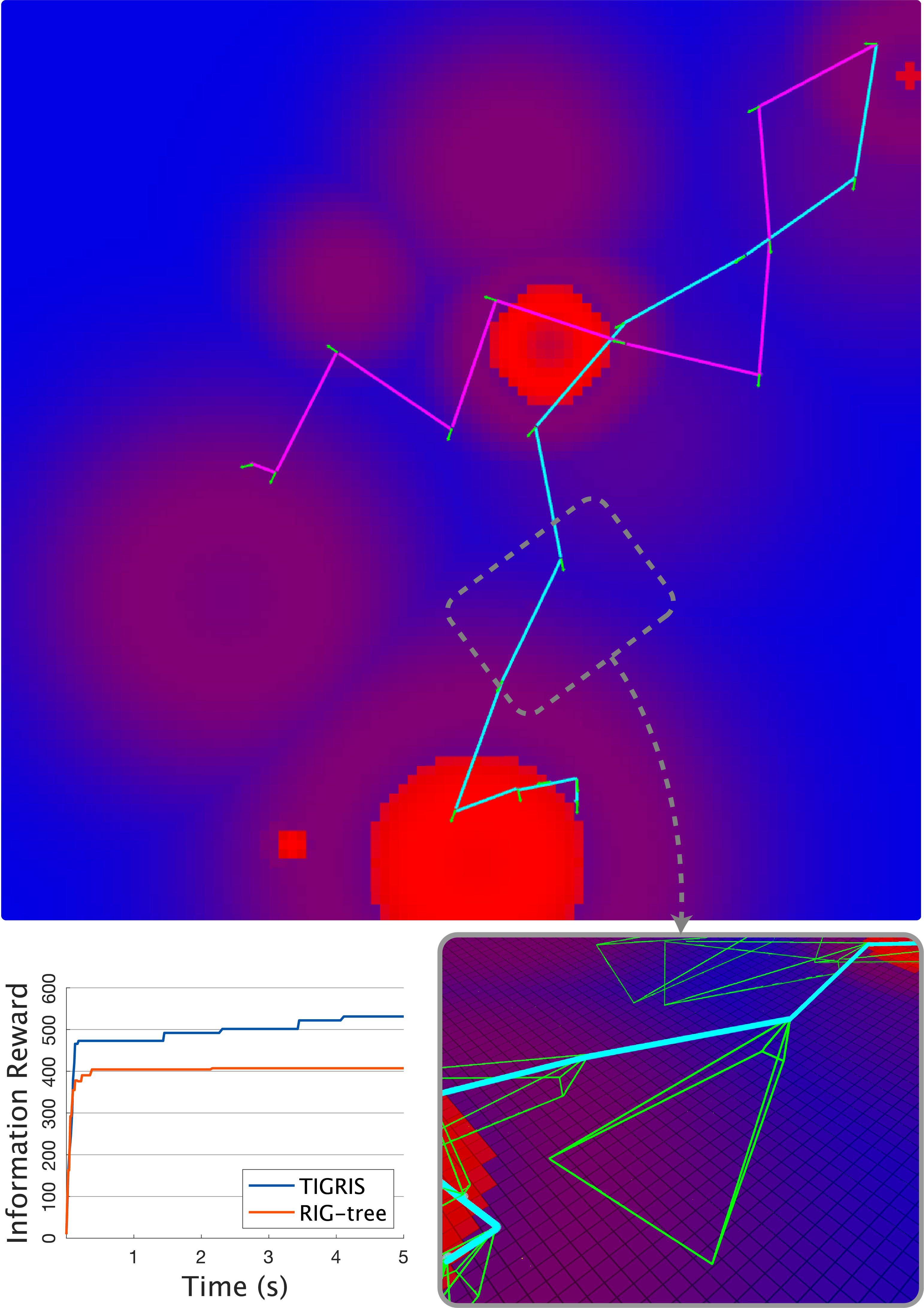}\caption{Example results in the testing environment. Red regions are areas of high reward and blue is low reward. The blue lines is the TIGRIS solution and the pink line is RIG-tree. The bottom right image shows a 3D view of the environment, and the bottom left graph shows the best path reward of the planners over time.}
    \label{fig:example_results}        
\end{figure}

Fig. \ref{fig:example_results} shows one result within the simulation environment. The planned path by TIGRIS is seen in blue and has an information reward of $531.2$ while the RIG-tree path is in pink and has an information reward of $407.2$. The green arrows at each node show the $\psi$. The background color show the information reward for each cell with blue being low reward and red being high. The image on the bottom right shows the 3D view of the environment and helps visualize the viewing frustum and sensor footprint. From looking at the results, you can see how TIGRIS maximizes the amount of information reward with the given budget by viewing all of the highest reward regions, even multiple times. 


It is also interesting to note the value of the best trajectory reward over time. 
Because the baseline did not include edge rewards, it would often miscalculate which trajectory was the best. Often a worse path would be selected over a better path because an area of high reward under an edge was not accounted for. A graph of the best path information reward versus time can be seen in the bottom left of Fig. \ref{fig:example_results}. The best path reward for the baseline decreased at some points in the beginning of the planning. Even with the additional computation time due to edge reward calculations, testing showed TIGRIS found better paths by using our edge information reward approximation. Another benefit to adding edge rewards is it allows for larger extend distances while still being able to closely approximate the information reward over the trajectory. This means TIGRIS can plan in large spaces with a less dense tree than the baseline. Without the edge rewards, calculating the trajectory reward with only the nodes would be a poor approximation of the path reward and would not reflect the reality that information is also gathered between the nodes.


\section{Conclusion} \label{sec:conclusion}
This work presents a novel informed sampling-based algorithm for informative planning called TIGRIS. Our method is tested using Monte Carlo simulations and is shown to have a statistically significant improvement over a comparable baseline. TIGRIS does well at quickly finding paths that maximize the information reward and reducing entropy over the belief space. As the budget increases to be large relative to the search space, the performance of TIGRIS would be expected to degrade as the problem approaches that of a coverage planner and the density of the tree would need to be intractably high. In these situations, replanning on the fly with TIGRIS could be a potential solution and is slated for future work. Additional future directions of this work could include searching for and tracking moving target, local optimization of the final path, or local optimization over path segments within the planning framework.






\bibliography{ref}

\end{document}